**Deep Learning-Driven Heat Map Analysis for Evaluating thickness of Wounded Skin Layers**

Devakumar GR, JB Kaarthikeyan, Dominic Immanuel T, Sheena Christabel Pravin*

School of Electronics Engineering,

Vellore Institute of Technology, Chennai.

*sheenachristabel.p@vit.ac.in

**Abstract**

Understanding the appropriate skin layer thickness in wounded sites is an important tool to move forward on wound healing practices and treatment protocols. Methods to measure depth often are invasive and less specific. This paper introduces a novel method that is non-invasive with deep learning techniques using classifying of skin layers that helps in measurement of wound depth through heatmap analysis. A set of approximately 200 labeled images of skin allows five classes to be distinguished: scars, wounds, and healthy skin, among others. Each image has annotated key layers, namely the *stratum cornetum*, the *epidermis*, and the *dermis*, in the software Roboflow. In the preliminary stage, the Heatmap generator VGG16 was used to enhance the visibility of tissue layers, based upon which their annotated images were used to train ResNet18 with early stopping techniques. It ended up at a very high accuracy rate of 97.67%. To do this, the comparison of the models ResNet18, VGG16, DenseNet121, and EfficientNet has been done where both EfficientNet and ResNet18 have attained accuracy rates of almost 95.35%.

For further hyperparameter tuning, EfficientNet and ResNet18 were trained at six different learning rates to determine the best model configuration. It has been noted that the accuracy has huge variations with different learning rates. In the case of EfficientNet, the maximum achievable accuracy was 95.35% at the rate of 0.0001. The same was true for ResNet18, which also attained its peak value of 95.35% at the same rate. These facts indicate that the model can be applied and utilized in actual-time, non-invasive wound assessment, which holds a great promise to improve clinical diagnosis and treatment planning.

1. **Introduction**

It has emerged as a huge phenomenon of deep learning for medical image analysis, especially in wound analysis and tissue monitoring, with non-invasive accuracy and interpretability that makes approaches in clinical applications critical. Traditional methods involved have always been invasive risks while it becomes highly impractical, such as real-time and individualized care in sensitive cases, like post-surgical monitoring, burn care, diabetic wound management, and recovery after cancer treatment [1][2]. Continuous monitoring of wound healing with minimal patient discomfort has become a crying necessity henceforth, and especially in cases that demand detailed skin layer analysis, as seen in reconstructive surgeries related to breast cancer patients where healing dynamics could be extremely complex and therefore must be monitored accurately.

In the following paper, we discuss how all these problems are overcome with a deep learning pipeline. We used a dataset of 200 annotated skin tissue images into five different categories: *"scar1", "scar2", "wound", "healthy_men", and "healthy_women"*. VGG16 would be applied to create heatmaps of the OCT images of skin layers that have improved interpretability, with the most important layers of the skin being lit. These heatmaps are then manually annotated into three layers of human skin: *stratum corneum*, *epidermis*, and *dermis*, which we did using Roboflow. We first trained a ResNet18 model to classify these annotated layers, early-stopped to prevent overtraining, which actually culminated with an impressive accuracy of 97.67%, and the classification report illustrates high precision and recall across all classes.

Further, we performed an ablation study by making early stopping absent in ResNet18, VGG16, DenseNet121, and EfficientNet models. The results obtained show that optimal accuracies were reached with both the ResNet18 and EfficientNet models at 95.35%, while that of DenseNet121 was at 90.70%, and that of VGG16 only at 72.09% accuracy. This therefore proves the strength of ResNet18 in our task. Finally, for optimizing the two models, EfficientNet and ResNet18, we selected six values of learning rates; we concluded that for both models, 0.0001 would be the best learning rate that could sustain high accuracies ranging to 95%.

In recent years, several studies have surfaced toward potential applicability and challenges in the application of deep learning in wound analysis. For instance, Shenoy et al. [3] applied ensemble CNNs in the DeepWound model in monitoring postoperative wounds using images captured by smartphones while attaining good accuracy despite the limitations of the dataset. He et al. [4] combined a traditional with a LayerCAM-based approach to analyze the collagen fibers in the histological wound images, which further rendered interpretability but was limited by a small dataset. Blanco et al. [5] proposed a superpixel-driven ResNet model for dermatological ulcers with excellent accuracy but bears a high computational requirement. Zhang et al. [6] and Siregar et al. [7] also proved the multifaceted and fine resolutions of deep learning but limitations in computation resources and dataset sizes. Zhao and Chen [8] studied hydrogel-based wound healing using Optical Coherence Tomography, and Neto et al. [9] developed a predictive model for wound state with very cheap sensors—high promise for real-time clinical applicability.

This approach leverages deep learning models, coupled with the mechanisms of heatmap interpretability along with a robust classification mechanism, which overcomes shortcomings in providing an accurate clinical tool for wound analysis. It is a pipeline that integrates CNNs and heatmaps outlining why interpretability in medical AI and non-invasive evaluation are critically important for widespread clinical adoption.

## 2. Related work

Significant advancements in wound assessment and monitoring using deep learning have demonstrated the potential of AI-driven techniques in medical diagnostics. Shenoy et al. (2018) introduced "DeepWound," an automated wound assessment tool leveraging convolutional neural networks (CNNs) and VGG16-based models to classify postoperative wounds. DeepWound achieved high accuracy using CLAHE preprocessing and data augmentation, and a mobile app (Theia) facilitated real-time monitoring. However, this approach faced challenges with privacy and dataset limitations [1].

He et al. (2020) focused on wound healing progress by analyzing collagen fiber regions in histological images using a fine-tuned VGG16 model with LayerCAM interpretability. This approach successfully combined deep learning with traditional statistical analyses, achieving 82% accuracy in wound stage classification. However, the complexity of LayerCAM limited generalizability due to a small dataset size [2].

Blanco et al. (2019) developed QTDU, a ResNet-based superpixel-driven approach for dermatological wounds, which segmented and quantified ulcer areas with an AUC of 0.986. QTDU's precise tissue quantification demonstrates its potential in dermatological wound analysis, though its reliance on expert-labeled images limits scalability [3].

Zhang et al. (2022) conducted a comprehensive survey on wound image analysis, implementing ResNet101 with a learning rate scheduler on a breast cancer wound dataset. This study achieved an accuracy of 97% but noted challenges like computational costs and model overfitting [4].

Siregar et al. (2024) used U-Net and Mask R-CNN models to measure wound size in zebrafish, an ethical and cost-effective alternative to rodent models. However, the transition to mammalian models posed challenges [5]. Zhao and Chen (2023) integrated hydrogels in skin wound healing, using Optical Coherence Tomography (OCT) and deep learning to assess hydrogel performance. Although effective in wound healing, hydrogel preparation complexity and cytotoxicity remain limitations [6].

Lastly, Ribeiro Neto (2021) presented a cost-effective, sensor-based wound healing tracking model, achieving an 85.7% accuracy rate. The model demonstrated potential for real-time monitoring, although further refinement is needed for clinical applications [7].

Building on these approaches, our study integrates heatmap-based VGG16 for image preprocessing and ResNet18 and EfficientNet for classification, employing early stopping and learning rate optimization to improve accuracy and reduce overfitting. This model not only achieves high accuracy in wound classification and skin layer analysis but also provides a scalable and interpretable tool for real-time wound monitoring, addressing limitations in previous studies regarding data availability, computational efficiency, and generalizability.

### 3. Proposed Framework

This proposed framework initiated with about 200 images of skin tissue, each belonging to five classes: *"scar1", "scar2", "wound", "healthy_men", and "healthy_women"*. Better vision for those features important for classification is achievable through using VGG16 first to transform the images into heatmaps. The heatmaps provide detailed visualizations of skin tissue variations, making it clearer to identify tissue structures for further analysis.

The images were then imported into Roboflow for manual annotation on three important layers of the skin, namely *stratum cornetum* and *epidermis* and *dermis*.

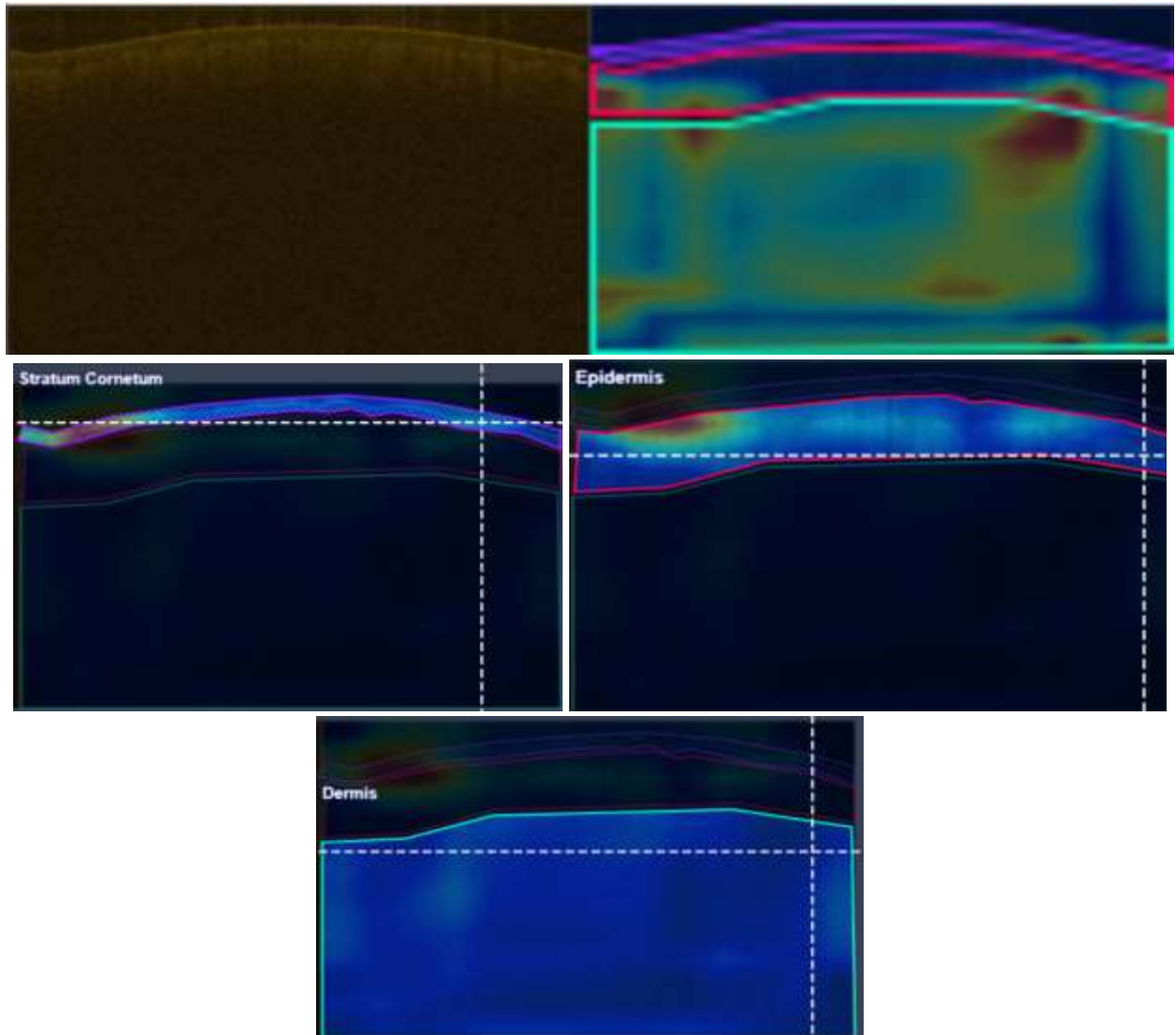

Fig. 1: Generated Heatmap and Annotations of the different layers

The annotations were necessary to take apart small structural differences in the tissue needed to raise the accuracy of classification for the model. The dataset itself was well-structured under separate folders for images and labels of the YOLOv8 format.

Different architectures of deep learning for model development, such as VGG16, DenseNet121, and EfficientNet have been researched. Comparatively, the best model to apply to this problem is the ResNet18 model, which has a unique residual learning mechanism to overcome the problems resulting from vanishing gradients, a very common problem in deep neural networks. Skip connections" or shortcuts between layers are the crux of residual networks. This facilitates passing the input of one layer directly to the output of a deeper layer so that the network may learn more complex representations without degradation of performance.

Mathematically, the residual learning framework can be written as follows:

$$y = F(x, \{Wi\}) + x$$

where x represents the input, F the residual mapping function that was learned by the layer and Wi the weights in the layer. This helps the network learn the residual function instead of trying directly to learn some complicated transformation, thereby making training for deep networks more efficient.

ResNet18 is composed of 18 layers arranged in blocks, where each block contains two convolutional layers with batch normalization and ReLU activation between them and followed by a skip connection. It's this architecture that enables the preservation of integrity as well as features throughout the layers. So, it would make it possible to have more stable and efficient training even with deeper networks. So, ResNet18 also maintained high accuracy for our dataset as well, achieving a final accuracy of 97.67% with early stopping to prevent overfitting.

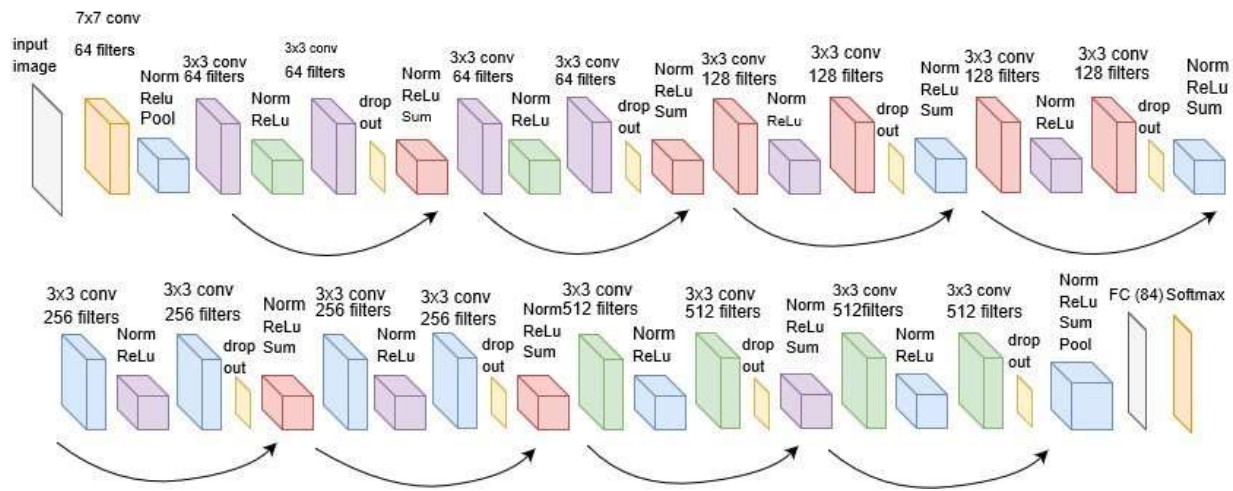

Fig. 2: Resnet18 Architecture

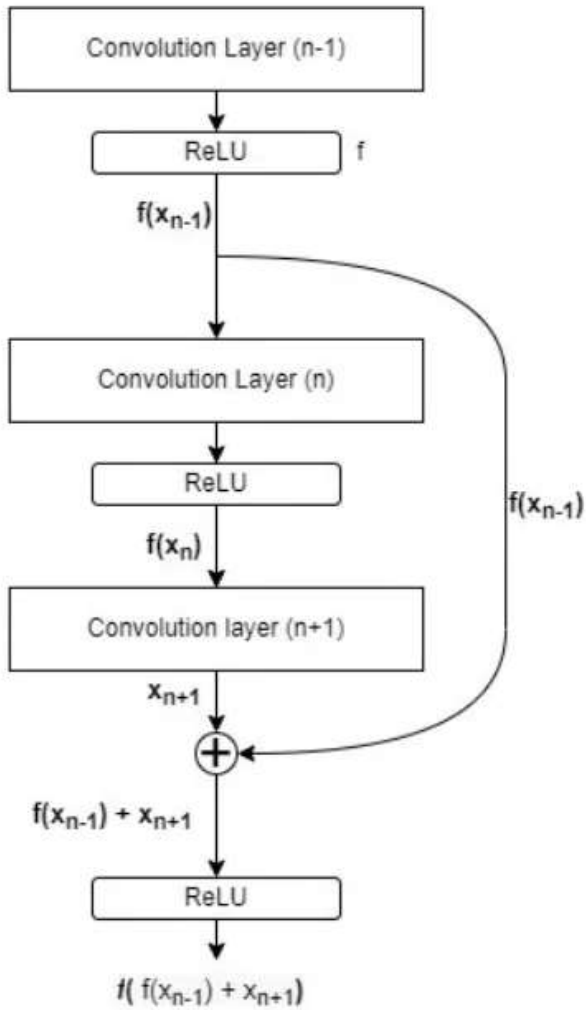

Fig. 3: Model Flowchart

| layer name | output size | 18-layer | 34-layer | 50-layer | 101-layer | 152-layer |
|---|---|---|---|---|---|---|
| conv1 | 112×112 | | 7×7, 64, stride 2 | | | |
| | | | 3×3 max pool, stride 2 | | | |
| conv2_x | 56×56 | $\begin{bmatrix} 3\times3, 64 \\ 3\times3, 64 \end{bmatrix} \times 2$ | $\begin{bmatrix} 3\times3, 64 \\ 3\times3, 64 \end{bmatrix} \times 3$ | $\begin{bmatrix} 1\times1, 64 \\ 3\times3, 64 \\ 1\times1, 256 \end{bmatrix} \times 3$ | $\begin{bmatrix} 1\times1, 64 \\ 3\times3, 64 \\ 1\times1, 256 \end{bmatrix} \times 3$ | $\begin{bmatrix} 1\times1, 64 \\ 3\times3, 64 \\ 1\times1, 256 \end{bmatrix} \times 3$ |
| conv3_x | 28×28 | $\begin{bmatrix} 3\times3, 128 \\ 3\times3, 128 \end{bmatrix} \times 2$ | $\begin{bmatrix} 3\times3, 128 \\ 3\times3, 128 \end{bmatrix} \times 4$ | $\begin{bmatrix} 1\times1, 128 \\ 3\times3, 128 \\ 1\times1, 512 \end{bmatrix} \times 4$ | $\begin{bmatrix} 1\times1, 128 \\ 3\times3, 128 \\ 1\times1, 512 \end{bmatrix} \times 4$ | $\begin{bmatrix} 1\times1, 128 \\ 3\times3, 128 \\ 1\times1, 512 \end{bmatrix} \times 8$ |
| conv4_x | 14×14 | $\begin{bmatrix} 3\times3, 256 \\ 3\times3, 256 \end{bmatrix} \times 2$ | $\begin{bmatrix} 3\times3, 256 \\ 3\times3, 256 \end{bmatrix} \times 6$ | $\begin{bmatrix} 1\times1, 256 \\ 3\times3, 256 \\ 1\times1, 1024 \end{bmatrix} \times 6$ | $\begin{bmatrix} 1\times1, 256 \\ 3\times3, 256 \\ 1\times1, 1024 \end{bmatrix} \times 23$ | $\begin{bmatrix} 1\times1, 256 \\ 3\times3, 256 \\ 1\times1, 1024 \end{bmatrix} \times 36$ |
| conv5_x | 7×7 | $\begin{bmatrix} 3\times3, 512 \\ 3\times3, 512 \end{bmatrix} \times 2$ | $\begin{bmatrix} 3\times3, 512 \\ 3\times3, 512 \end{bmatrix} \times 3$ | $\begin{bmatrix} 1\times1, 512 \\ 3\times3, 512 \\ 1\times1, 2048 \end{bmatrix} \times 3$ | $\begin{bmatrix} 1\times1, 512 \\ 3\times3, 512 \\ 1\times1, 2048 \end{bmatrix} \times 3$ | $\begin{bmatrix} 1\times1, 512 \\ 3\times3, 512 \\ 1\times1, 2048 \end{bmatrix} \times 3$ |
| | 1×1 | | average pool, 1000-d fc, softmax | | | |
| FLOPs | | $1.8\times10^9$ | $3.6\times10^9$ | $3.8\times10^9$ | $7.6\times10^9$ | $11.3\times10^9$ |

Fig. 4: Resnet architecture review table

The early stopping monitors the validation loss so that training is stopped when improvements become stagnant for a certain patience interval. So, this moved ResNet18 to further great heights by avoiding overfitting in the annotated dataset. This, combined with optimizing the learning rate, helped in tuning the ability of the model to generalize effectively on unseen data, and the trained model retained high accuracy across different tissue types and layers for our objectives. Overall, the proposed framework here highlights those architectures like ResNet18, particularly with a residual learning mechanism, are effective in robust classification of skin tissues. With regard to our structured preprocessing combined with accurate annotation, the systemic approach used to train and test our framework gives great promise for scalable assessment of skin tissue, opening up possible future developments on self-aware wound analysis.

### 4. Results and Discussions

This section summarizes the results of training, optimizing, and evaluation of the ResNet18 architecture model for the classification of skin tissue images. The work discusses how architecture impacts a choice for a model; annotations impact interactions of models; impacts of methodologies applied during the training process; and an analysis of errors.

### 4.1 Overview of the ResNet18 Model

ResNet18 is one of the most advanced architectures of convolutional neural networks (CNN) with residual learning using skip connections; this, in effect, allows gradients to bypass certain layers and hence mitigates the vanishing gradient problem to make deeper learning possible with relatively lesser computation. The 18-layer architecture includes convolutional, pooling, and fully connected layers to extract hierarchical features for useful applications by employing them in complicated datasets. Due to the lightweight architecture, it is computationally efficient in tasks demanding high accuracy, such as in medical image analysis.

### 4.2 Annotations Impact

The input of pre-labeled labels from Roboflow had a highly positive impact on the quality of this dataset. Every image was delineated into three well-defined layers of *stratum cornetum*, *epidermis*, and *dermis* as bounding boxes. This provided the model with the additional data above the original images, thus allowing the model to focus completely on region-based features. This greatly improved the precision for classification, and it stands to bear testament to the role of proper labeling in a machine learning workflow.

### 4.3 Training and Optimization

ResNet18 is trained using an 80-20 split of the dataset between training and test sets. The initial learning rate is 0.001, while the optimizer used is Adam, which ultimately provided good gradient updates with the categorical cross-entropy loss function.

$$loss(x, class) = -log(\frac{exp(x[class])}{\Sigma j\ exp(x[j])})$$

An early stopping strategy was implemented with patience of 10 epochs where the validation loss is controlled in which training is stopped when it undergoes stagnation. This not only halts overfitting but also maximizes training efficiency.

The training stage consisted of 100 epochs at maximum. Dropout was used between each layer so that the layers were exposed to reducing overfitting by randomly deactivating 50% of neurons per epoch. Training, validation loss, and accuracy are tracked and plotted as functions of epochs.

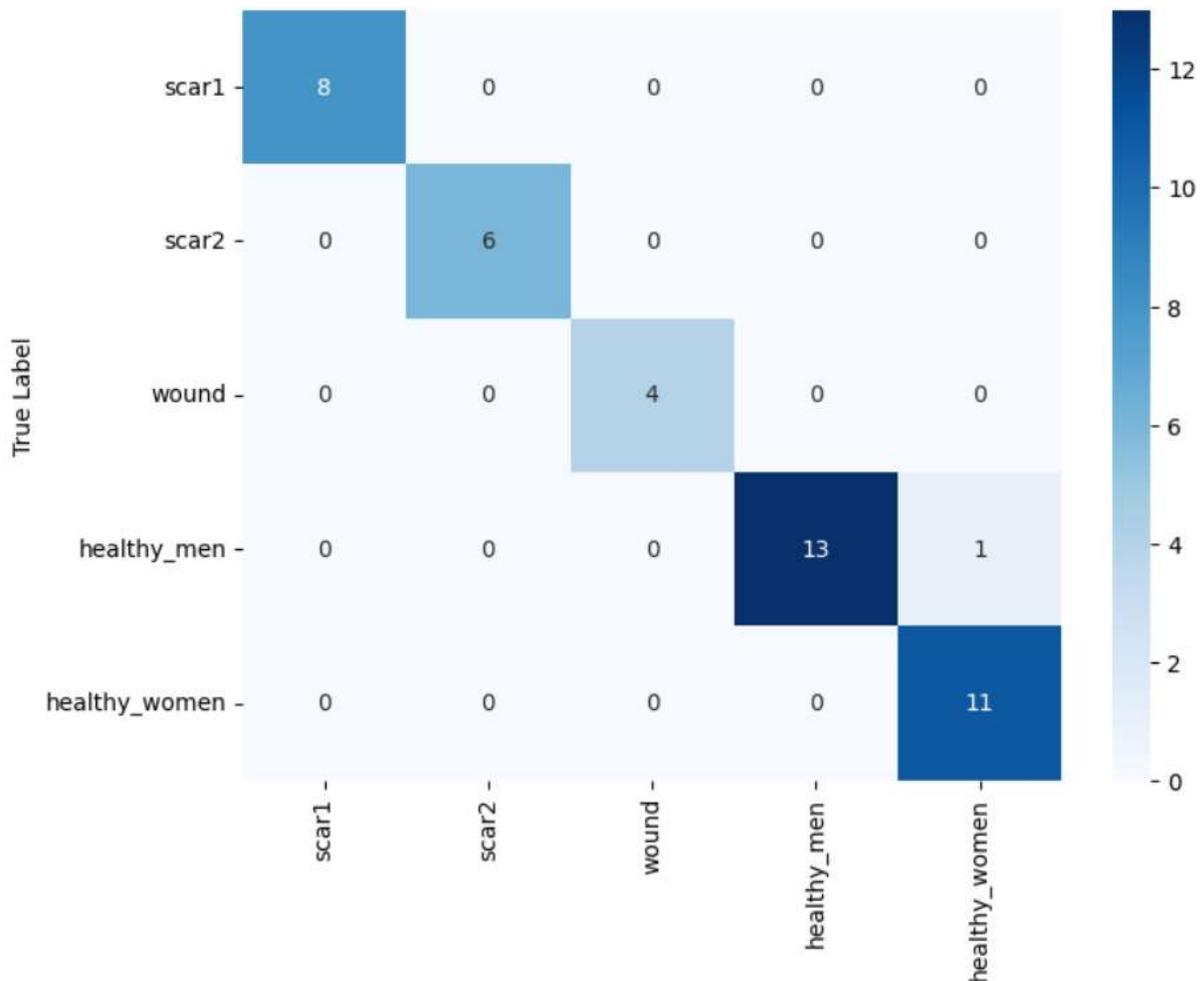

### 4.4 Model Performance
On the test set, ResNet18 accuracy reached an impressive 97.67% accuracy rate. Ablation studies showed that the architectures outperform other variants such as VGG16, DenseNet121, and EfficientNet. Other metrics included precision, recall, F1-score, and a confusion matrix in the form of the following formulas:

$$\text{Accuracy} = \frac{TP + TN}{TP + TN + FP + FN}$$

$$\text{Precision} = \frac{TP}{TP + FP}$$

$$\text{Recall} = \frac{TP}{TP + FN}$$

$$\text{F1-Score} = 2 \times \frac{\text{Precision} \times \text{Recall}}{\text{Precision} + \text{Recall}}$$

It turns out that the confusion matrix reveals high accuracy in classification for all classes, and only minor misclassifications between healthy_men and healthy_women might be due to overlapping features.

**4.5 Error Analysis**

As it turns out, similar visual features between healthy_men and healthy_women were initial causes for failure to differentiate between these classes. Including them in the dataset with the Roboflow annotations generally fixed this problem and permitted the model to rely on deciding key characteristics differentiating classes. Aiding with the addition of the annotations as auxiliary parameters for input contributed to a noticeable gain in class-specific recall and overall accuracy.

These enhancements thus appear to suggest that if the labeling is exact, region-specific information is important for augmenting the performance of the model, especially in tasks requiring fine visual differences.

Also, the graph shows a gradually descending line, which shows that the model training is smooth with a gradual loss drop, which shows no overfitting has taken place. In the upcoming section, the graphs are very spiky, which shows how important triggers are in the proposed model.

## 5. Ablation Study

In the ablation study, four distinct pre-trained models: ResNet18, VGG16, DenseNet121, and EfficientNet, were used to see how these models performed without early-stopping triggers. Since the goal was assessing how well the models performed on the task of classification of skin tissue by making use of annotations from Roboflow to enhance the general learning process, this experiment was important with labeled data from Roboflow because it allows for better understanding of characteristics of each class, hence better classification accuracy.

The test set ResNet18, famous for its efficient residual learning framework, achieved 95.35% accuracy in the classification of skin tissue images. Due to the removal of the vanishing gradient problem, the model architecture including residual connections enables it to learn better. These help gradients flow during backpropagation easier, hence a robust model for the classification of images, ResNet18. Even if early stopping triggers were removed, the accuracy would have dropped by roughly about 2%.

The second most widely used model, VGG16, performed poorly with an accuracy of 72.09%. Its architecture is based on simple yet deep networks of convolutional layers but is infamous for having large numbers of parameters that can make it sometimes prone to overfitting compared to some complex architectures like ResNet. This overfitting was strongly reflected in the relatively lower accuracy that VGG16 achieved, especially when the training was performed without early stopping triggers that might have helped in preventing it.

The DenseNet121 model, which connects each layer to every other layer in a dense block, shows reasonably good performance at 90.70%. The architecture of DenseNet generally has a better gradient flow and therefore promotes feature reuse across the different layers. Therefore, it can be very useful when dealing with very complex datasets. Still, it is not as effective as ResNet18 and EfficientNet, based probably on its more complex architecture, and it may need to undergo more fine-tuning to achieve its perfect performance.

The model EfficientNet scaled with both depth, width, and resolution achieved accuracy similar to that of ResNet18 at 95.35%. With the compound scaling approach, EfficientNet ensured the use of as few parameters as possible in the model, which made it one of the best models used in this study. It performed well in the beginning, but when the step of early stopping was removed, the results showed that the model needed more careful optimizations for consistent high performance.

Based on the above initial results, the best among the two models is ResNet18 and EfficientNet. We selected these two models and further trained them with different learning rates for fine-tuning the performance.

The learning rate values from very low to high are set as follows for both ResNet18 and EfficientNet that is [1e-5, 1e-4, 1e-3, 1e-2, 1e-1, 1]. For EfficientNet, the accuracy was 95.35%, which was identical with that of the first training, at a learning rate of 0.0001. The accuracy plunged sharply to as low as remarkably 4.65% when the learning rate was 0.1 and down to 30.23% at 1.0.

For ResNet18, the optimal results were also obtained using a learning rate of 0.0001, at which the accuracy has been reached at 95.35%. However, if a higher learning rate is used, then the model's result is significantly poor. At learning rates of 0.01, 0.1, and 1.0, the accuracy has been reduced to 34.88%, meaning that learning rates at such high levels prevented the model from converging suitably.

The study demonstrates, through results, that learning rate selection is one of the key factors for the successful training of models. Notably, both ResNet18 and EfficientNet attained very high accuracy at lower learning rates, but very high values of learning rates led to serious loss in accuracy, which does highlight the need for fine-tuning hyperparameters for the best achievable performance.

Besides that, the accuracy results that have been attained for the models along with the corresponding training graphs and confusion matrices further illustrate how hyperparameter tuning, including learning rate adjustment, heavily weighs on the models' performances. Annotations from Roboflow also came in handy, proving to be important for achieving better accuracy by giving labeled data that was used as input to train the models concerned.

ResNet18:

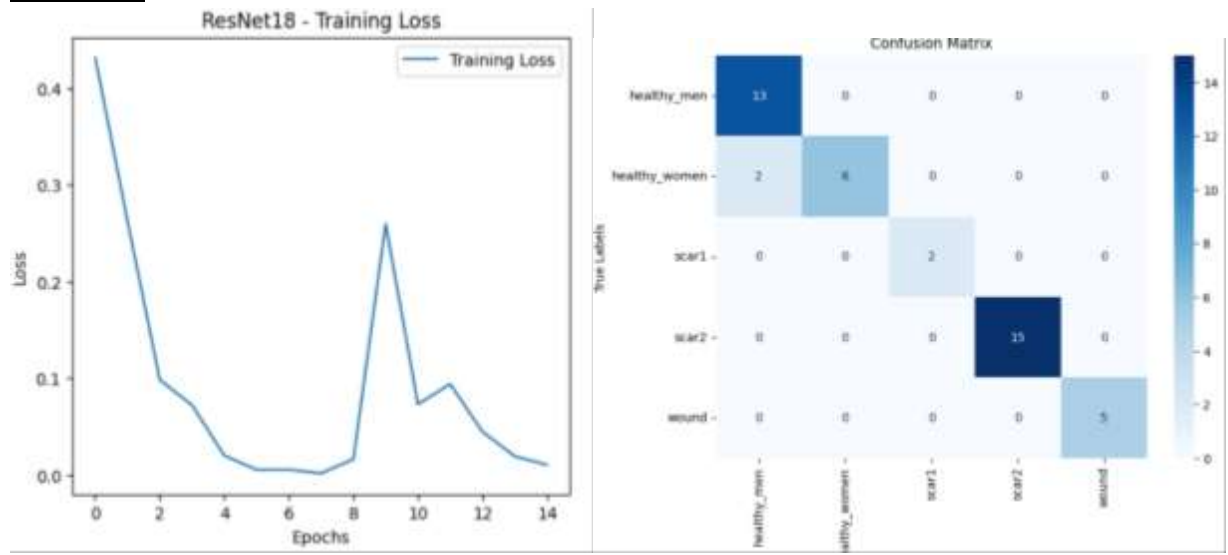

Fig. 7: Results obtained using ResNet

VGG16:

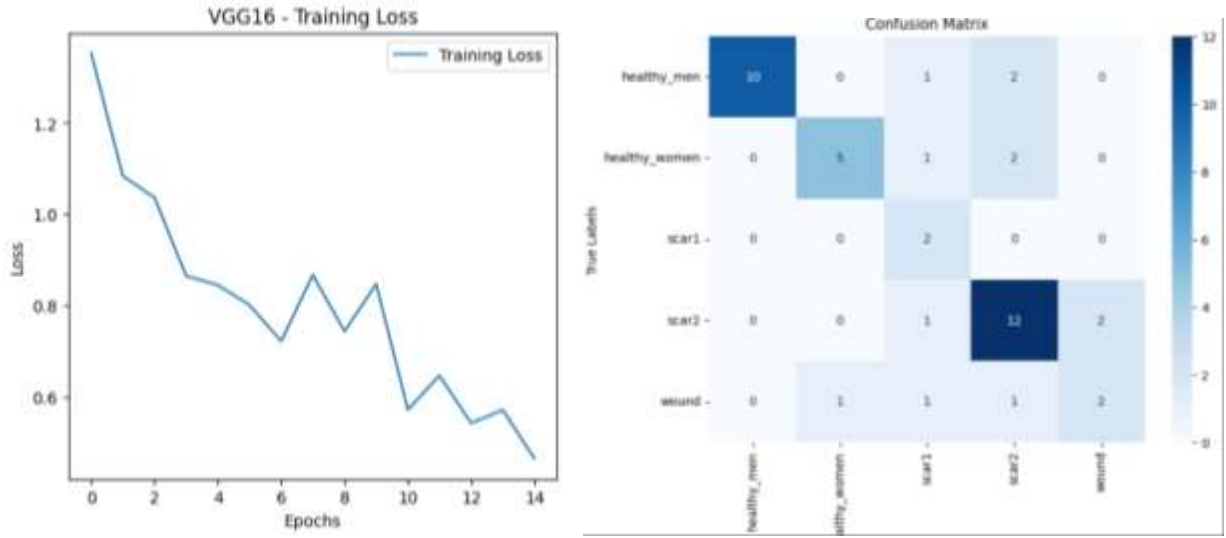

Fig. 8: Results obtained using VGG16

DenseNet121:

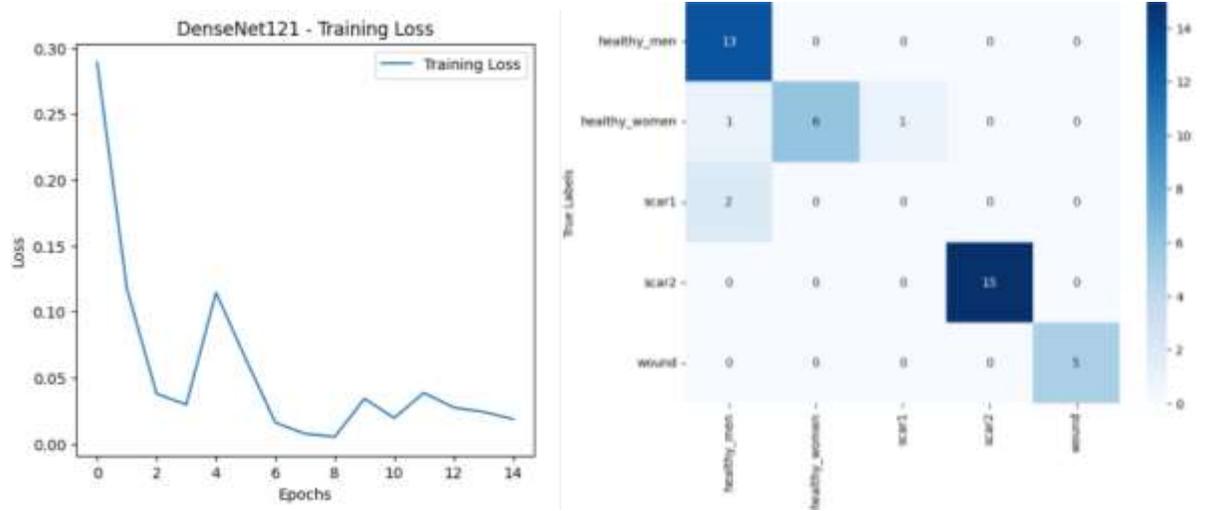

Fig. 9: Results obtained using Densenet121

EfficientNet:

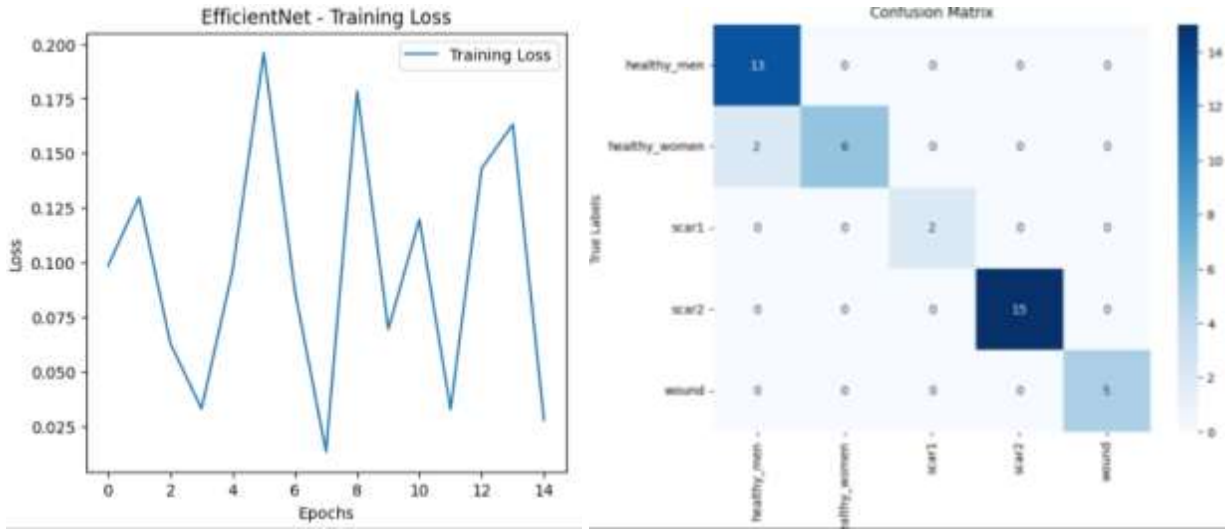

Fig. 10: Results obtained using Efficient net

This ablation study thereby brings out the substantial impact that model architecture choices and hyperparameter optimization would have on the final performance, reiterating the notion that precise tuning and quality data annotations are essential.

## 6. Conclusion

In this paper, we shall discuss and test the performance of four renowned deep learning models: ResNet18, VGG16, DenseNet121, and EfficientNet on the classification of images of skin tissues into five classes. We did not apply early stopping and have an ablation study on the optimization of some hyperparameters to keep optimal learning rates. In this research, the top two performers included ResNet18 and EfficientNet with an accuracy of 95.35%.

The best use of the labeled annotations from Roboflow was to make the models understand nuances in classes of skin tissue. Further analysis on how the models perform with different learning rates shows that fine-tuning the hyperparameter is essential for best performance. Both ResNet18 and EfficientNet showed the best result when fine-tuned at a learning rate of 0.0001, proving the need for a more precise fine-tuning factor to avoid overfitting or underfitting.

From the results presented in this study, it can be concluded that the selection of model, effects of hyperparameter tuning, and quality labeled data are involved in training deep learning models based on an image classification task. So, other further work may involve designing techniques such as transfer learning with domain-specific models, architectural changes, and higher optimization of learning rates to improve the classification accuracy even further.